\documentclass[pmlr]{jmlr}

\usepackage{longtable}
\usepackage{cleveref}
\usepackage{xcolor}
\usepackage{comment}
\definecolor{rulescolor}{RGB}{200,200,200}
\usepackage{booktabs}
\usepackage{wrapfig,lipsum}
\usepackage{multirow}
\usepackage[load-configurations=version-1]{siunitx} 

\theorembodyfont{\upshape}
\theoremheaderfont{\scshape}
\theorempostheader{:}
\theoremsep{\newline}

\jmlrvolume{1}
\jmlryear{2023}
\jmlrworkshop{}

\title[A Retrospective of the MineRL BASALT 2022 Competition]{Towards Solving Fuzzy Tasks with Human Feedback: \titlebreak A Retrospective of the MineRL BASALT 2022 Competition}

  \author{\Name{Stephanie Milani} \Email{smilani@cs.cmu.edu}\\
  \addr Carnegie Mellon University
  \AND
  \Name{Anssi Kanervisto}\thanks{Equal contribution} \Email{anssi.kanervisto@microsoft.com}\\
  \addr Microsoft Research, Cambridge 
  \AND 
  \Name{Karolis Ramanauskas$^*$} 
  \Email{kr711@bath.ac.uk}\\
  \addr University of Bath
  \AND 
  \Name{Sander Schulhoff} 
  \Email{sschulho@umd.edu}\\ 
  \addr University of Maryland
  \AND 
  \Name{Brandon Houghton} 
  \Email{brandon@openai.com}\\ 
  \addr OpenAI
  \AND 
  \Name{Sharada Mohanty} 
  \Email{mohanty@aicrowd.com}\\
  \addr AIcrowd
  \AND 
  \Name{Byron Galbraith} 
  \Email{}
  \addr 
  \AND 
  \Name{Ke Chen} 
  \Email{chenke3@corp.netease.com}\\
  \Name{Yan Song}
  \Email{songyan@corp.netease.com}\\
  \Name{Tianze Zhou}  
  \Email{zhoutianze@corp.netease.com}\\
  \Name{Bingquan Yu}
  \Email{yubingquan@corp.netease.com}\\
  \Name{He Liu}
  \Email{liuhe1@corp.netease.com}\\
  \Name{Kai Guan} 
  \Email{guankai1@corp.netease.com|}\\
  \Name{Yujing Hu} 
  \Email{huyujing@corp.netease.com}\\
  \Name{Tangjie Lv}
  \Email{hzlvtangjie@corp.netease.com}\\ 
  \addr NetEase Fuxi AI Lab 
  \AND 
  \Name{Federico Malato}
  \Email{fmalato@uef.fi}\\
  \addr University of Eastern Finland
  \AND 
  \Name{Florian Leopold} 
  \Email{fleopold@techfak.uni-bielefeld.de}\\
  \addr University of Bielefeld
  \AND 
  \Name{Amogh Raut}
  \Email{}\\
  \addr Indian Institute of Technology 
  \AND 
  \Name{Ville Hautamäki}
  \Email{ville.hautamaki@uef.fi}\\
  \addr University of Eastern Finland 
  \AND 
  \Name{Andrew Melnik}
  \Email{andrew.melnik@uni-bielefeld.de}\\
  \addr University of Bielefeld
  \AND 
  \Name{Shu Ishida}
  \Email{ishida@robots.ox.ac.uk}\\
  \Name{João F. Henriques} 
  \Email{joao@robots.ox.ac.uk}\\
  \addr Visual Geometry Group, University of Oxford
  \AND 
  \Name{Robert Klassert} 
  \Email{robertklassert@pm.me}\\
  \addr Forschungszentrum Informatik Karlsruhe, Berkeley Existential Risk Initiative
  \AND 
  \Name{Walter Laurito} 
  \Email{lauritowal@yahoo.com} \\
  \addr Forschungszentrum Informatik Karlsruhe
  \AND 
  \Name{Ellen Novoseller}
  \Email{ellen.r.novoseller.ctr@army.mil}\\
  \Name{Vinicius G. Goecks} 
  \Email{vinicius.goecks@gmail.com}\\
  \Name{Nicholas Waytowich}
  \Email{nicholas.r.waytowich.civ@army.mil}\\
  \addr DEVCOM Army Research Laboratory 
  \AND 
  \Name{David Watkins}
  \Email{davidwatkins@cs.columbia.edu}\\
  \addr Boston Dynamics AI Institute, Columbia University
  \AND 
  \Name{Josh Miller}
  \Email{joshua.r.miller138.civ@army.mil}\\
  \addr DEVCOM Army Research Laboratory
  \AND 
  \Name{Rohin Shah} 
  \Email{rohinmshah@deepmind.com}\\ 
  \addr DeepMind
 }

\begin{document}

\maketitle
\newcommand{\todo}[1]{\textcolor{red}{TODO: #1}}
\newcommand{\smnote}[1]{\textcolor{orange}{#1}}
\newcommand{\smchange}[1]{\textcolor{black}{#1}}
\newcommand{\kr}[1]{\textcolor{black}{#1}}
\newcommand{\ak}[1]{\textcolor{black}{#1}}
\newcommand{\newchange}[1]{\textcolor{red}{#1}}

\newcommand{\namenospace}{MineRL \textsc{BASALT}}
\newcommand{\name}{MineRL \textsc{BASALT}\ }
\newcommand{\easy}{\textsc{intro}\ }
\newcommand{\fthf}{FTfHF}

\newcommand{\cavetasknospace}{\texttt{FindCave}}
\newcommand{\cavetask}{\cavetasknospace\ }
\newcommand{\waterfalltasknospace}{\texttt{MakeWaterfall}}
\newcommand{\waterfalltask}{\waterfalltasknospace\ }
\newcommand{\housetaskfull}{\texttt{BuildVillageHouse}\ }
\newcommand{\housetasknospace}{\texttt{House}}
\newcommand{\housetask}{\housetasknospace\ }
\newcommand{\pentaskfull}{\texttt{CreateVillageAnimalPen}\ }
\newcommand{\pentasknospace}{\texttt{AnimalPen}}
\newcommand{\pentask}{\texttt{AnimalPen}\ }
\newcommand{\mybox}[1]{\par\noindent\colorbox{rulescolor}
{\parbox{\dimexpr\textwidth-2\fboxsep\relax}{#1}}}
\begin{abstract}
\smchange{To facilitate research in the direction of fine-tuning foundation models from human feedback, we held the MineRL BASALT Competition on Fine-Tuning from Human Feedback at NeurIPS 2022. 
The BASALT challenge asks teams to compete to develop algorithms to solve tasks with hard-to-specify reward functions in Minecraft. 
Through this competition, we aimed to promote the development of algorithms that use human feedback as channels to learn the desired behavior. 
We describe the competition and provide an overview of the top solutions. 
We conclude by discussing the impact of the competition and future directions for improvement.}
\end{abstract}

\begin{keywords}
Learning from humans, fine-tuning, reward modeling, imitation learning, preference learning, reinforcement learning from human feedback
\end{keywords}

\newpage
\begin{figure}[t]
    \centering
    \includegraphics[trim=0 160 0 0,clip,width=.9\textwidth]{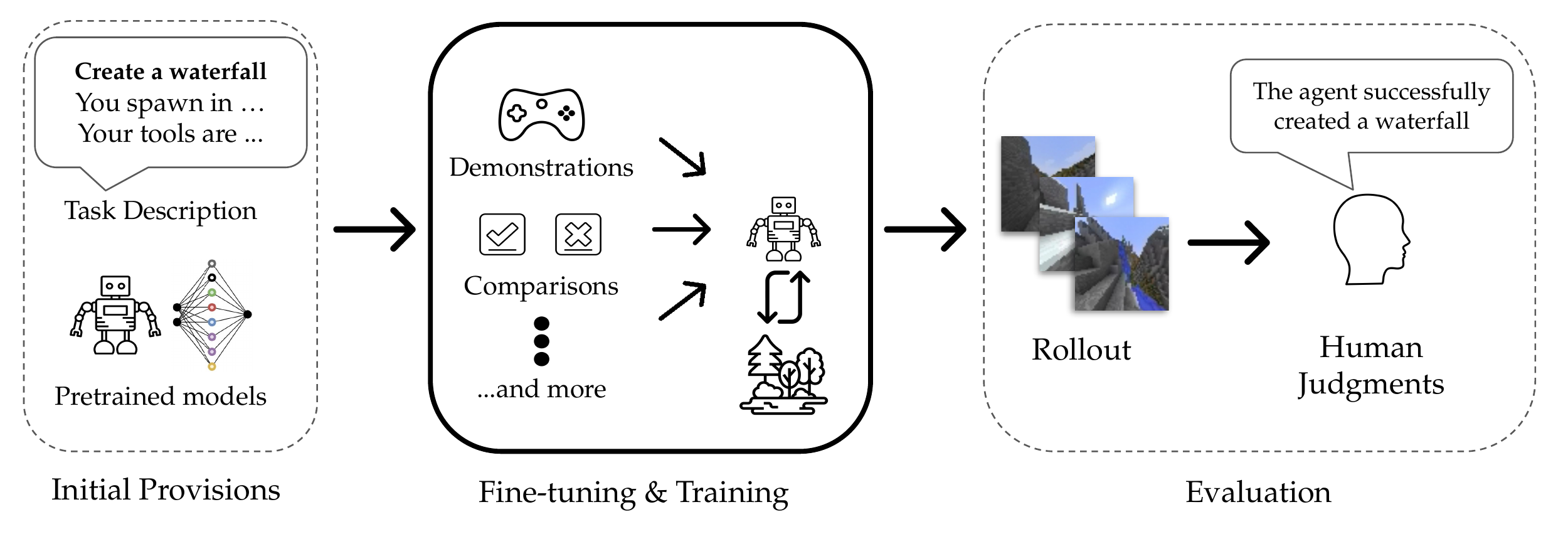}
    \caption{\textbf{The \name competition procedure.} \smchange{Participants developed their approaches through training and fine-tuning with human feedback. We evaluated the methods with human judgments and rank submissions according to their TrueSkill rating \citep{herbrich2006trueskill}.}\vspace{-8pt}}
    \label{fig:overview}
\end{figure}
\section{Introduction}
\label{sec:intro} 
\smchange{
The creation of foundation models~\citep{bommasani2021opportunities} has spurred advances across the field of machine learning, including few-shot learning~\citep{brown2020language}, sequential decision-making~\citep{carroll2022unimask}, and zero-shot image classification~\citep{radford2021learning}. 
In many tasks of real-world interest, exactly specifying an objective is challenging~\citep{kerr1975folly,russell2010artificial}.
Instead, these models are typically optimized to perform well with a self-supervised loss function, such as predicting the next word in an incomplete sentence~\citep{taylor1953cloze}, rather than the actual task of interest. 
To redirect these capabilities for the desired task, a natural idea is to then utilize human feedback for \textit{fine-tuning} these large pre-trained models. 
Although prior work has investigated the use of these feedback modalities for training~\citep{lin2020review}, few of these have been applied to foundation models outside of natural language applications~\citep{nakano2021webgpt,ouyang2022training}.}

\smchange{
To further encourage research progress in developing techniques for learning from human feedback for training foundation models, we held the second annual MineRL Benchmark for Agents that Solve Almost Lifelike Tasks (\namenospace) competition~\citep{shah2021minerl} at the 36th Conference on Neural Information Processing Systems (NeurIPS 2022). 
Different from last year, we focused on \textit{fine-tuning}.
To our knowledge, this competition is the first to test fine-tuning from human feedback (\fthf) techniques in the sequential decision-making setting. 
\Cref{fig:overview} provides a high-level overview of the competition.
Over the span of four months, teams developed agents for four challenging open-world tasks in Minecraft.
These tasks were specifically designed to have challenging task specifications, with hard-to-define reward functions that encourage a focus on \fthf.
More broadly, we see the impact of this competition as coming from the following sources.
The resulting techniques can enable the construction of AI systems for tasks without formal specifications and expand the set of properties that we can incorporate into such systems.
For more details about the broader impact of this competition, please see the competition proposal~\citep{shah2021minerl}.
In this paper, w}e present a high-level overview of the competition, a summary of the top solutions, and considerations for future competitions of this type.

\section{Competition Overview}
\label{sec:overview}

\smchange{We now present important details about the competition (more in \Cref{apd:comp_details}).} 

\paragraph{Tasks}
\smchange{
We provided a set of tasks in Minecraft that consist of a simple English-language task description and a Gym~\citep{brockman2016openai} environment. 
They were: \cavetasknospace, \waterfalltasknospace, \pentaskfull (\pentasknospace), and \housetaskfull (\housetasknospace).}
\smchange{
Crucially, the Gym environments lacked any associated reward functions.}
For each task, we provided a dataset of human demonstrations consisting of a sequence of state-action pairs.
\smchange{To help participants familiarize themselves with the code pipeline, we included an introductory track with a task with an explicit reward function, \texttt{ObtainDiamondShovel}.}

\paragraph{Resources}
\smchange{Through our partnership with AIcrowd, we provided competitors with a unified interface to register for the competition, submit trained agents, ask questions, and monitor their progress on a public leaderboard.}\footnote{\url{https://www.aicrowd.com/challenges/neurips-2022-minerl-basalt-competition}}
We partnered with OpenAI to release over 600h of human demonstrations in the four tasks and baselines built on the VPT model~\citep{baker2022video}. 
The baseline solution fine-tunes VPT model with behavioral cloning using the collected dataset.\footnote{The pre-trained models and human demonstration data are available from the following GitHub pages: \url{https://github.com/openai/Video-Pre-Training} and \url{https://github.com/minerllabs/basalt-2022-behavioural-cloning-baseline}.}
\ak{To provide mentorship and foster an active community, we continued maintaining the MineRL Discord server.} 

\paragraph{Rules and Validation}
\smchange{We required methods to use only the specified Gym API. 
For reproducibility, we mandated that participants submit their \textit{training} code.
We limited the size of data and public models (30MB) that could be included in their submissions. 
The provided resources did not count toward this limit. 
The AIcrowd page contains the full rules.\footnote{\url{https://www.aicrowd.com/challenges/neurips-2022-minerl-basalt-competition/challenge_rules}}}
\ak{
To ensure reproducibility, we retrained the finalists' submissions with their provided training code for up to 4 days (on all four tasks) using at most 10 hours of human feedback. 
We fixed the compute to 12 CPU cores, 56GB of RAM, and an NVIDIA Tesla K80 GPU. The training code could query humans for input with traditional desktop UIs. 
We provided human contractors who connected to the training instances to provide input. We also manually inspected the code to ensure compliance with the 30MB upload limit.
}

\paragraph{Evaluation}
\smchange{
\ak{Upon submission,} we deployed the agents on fixed world seeds of each task to generate multiple example videos.
\ak{After the submission deadline,} we asked human judges recruited through Amazon Mechanical Turk (MTurk) to choose which agent better completed the task through pairwise comparisons of the agents.
Given a dataset of these comparisons, we computed each agent's scores using the TrueSkill system \citep{herbrich2006trueskill}.
We determined the winners by normalizing and aggregating these scores across tasks.}

\paragraph{Prizes}
\ak{We awarded the top three solutions as ranked by the human evaluation 7000, 4000, and 3000 USD, respectively. 
\smchange{To encourage the exploration of creative solutions, we asked each advisor to select a single team to award a research prize of 1000 USD.} 
To drive community engagement, we gave a total of 1000 USD to participants who helped others or otherwise contributed to the competition.}

\paragraph{Related Competitions}
\smchange{
Minecraft is a popular platform for various research benchmarks~\citep{grbic2021evocraft,gray2019craftassist,johnson2016malmo,hafner2021benchmarking,fan2022minedojo} and competitions. 
Their diversity demonstrates Minecraft's flexibility to both instantiate and evaluate a variety of interesting problems. 
Research competitions have focused on multi-agent learning of cooperative and competitive tasks~\citep{perez2019multi}, generating functional and believable settlements~\citep{salge2018generative}, and learning to build structures based on natural language descriptions~\citep{kiseleva2021neurips}.
However, none of these competitions are particularly relevant to \fthf{} for hard-to-specify sequential decision-making tasks. 
The most related competitions are the previous MineRL Diamond competitions \smchange{that focused on} learning from a reward function and demonstrations to solve a crisply-defined task~\citep{guss2019neurips}. 
In contrast, we emphasize the use of fine-tuning techniques and utilize tasks that do not have an easy-to-define reward function.
}
\section{Solutions}
\label{sec:solutions}
\begin{table}[t]
    \centering
    \begin{tabular}{lcc}
    \toprule 
         & 2021 & 2022  \\
    \midrule
     Number of Teams & 37 & \textbf{63} \\
     Number of Individuals & 358 & \textbf{446} \\ 
     Number of Submissions & 271 & \textbf{504} \\
     Number of Teams that Scored Higher than Baselines & 10 & \textbf{11} \\
    \bottomrule
    \end{tabular}
    \caption{\textbf{Participation statistics.} The 2022 competition saw more teams, individuals, and submissions than 2021, demonstrating an increased interest.\vspace{-8pt}}
    \label{tab:basalt_participation}
\end{table}

\smchange{
Compared to last year, this year we saw a larger number of individuals (and teams) competing and nearly double the submissions (see \Cref{tab:basalt_participation}).
We believe that the number of submissions increased in part due to the introductory track. 
Interestingly, the scores corresponding to submissions that achieved higher than the lowest scores (e.g. approaches that did not just submit the baselines) remained the same as last year.
This similarity indicates that the increased popularity likely stemmed from less experienced teams.
}

\smchange{
In the rest of this section, we describe approaches taken by the competition winners and the teams who were selected for research prizes.\footnote{To see example videos of each of the agents trained using the algorithms described performing each of the tasks, please see: \url{https://www.youtube.com/playlist?list=PL7H6ODSaA0Y-yyJDXLOJQQcThg7_SBoqU}.}
}

\begin{table}[t]
\centering
\floatconts
  {tab:leaderboard}
  {\begin{tabular}{lSSSSS} \label{table:winners}
  \\ \toprule
  \bfseries Team & {\bfseries \cavetasknospace} & {\bfseries \waterfalltasknospace} & {\bfseries \pentasknospace} & {\bfseries \housetasknospace} & {\bfseries Average}\\
  \midrule
   GoUp & 0.31 & {\bfseries 1.21} & {\bfseries 0.28} & {\bfseries 1.11} & {\bfseries 0.73} \\
   UniTeam & {\bfseries 0.56} & -0.10 & 0.02 & 0.04 & {\bfseries 0.13} \\
   voggite & 0.21 & 0.43 & -0.20 & -0.18 & {\bfseries 0.06} \\
   JustATry & -0.31 & -0.02 & -0.15 & -0.14 & {\bfseries -0.15} \\
   TheRealMiners & 0.07 & -0.03 & -0.28 & -0.38 & {\bfseries -0.16} \\
   yamato.kataoka & -0.33 & -0.20 & -0.27 & -0.18 & {\bfseries -0.25} \\
   corianas & -0.05 & -0.26 & -0.45 & -0.24 & {\bfseries -0.25} \\
   Li\_and\_Ivan & -0.15 & -0.72 & -0.14 & -0.22 & {\bfseries -0.31} \\
   KAIROS & -0.35 & -0.32 & -0.41 & -0.36 & {\bfseries -0.36} \\
   Miner007 & -0.07 & -0.76 & -0.12 & -0.52 & {\bfseries -0.37} \\
   KABasalt & -0.57 & -0.23 & -0.41 & -0.31 & {\bfseries -0.38} \\
   \midrule
   Human2 & 2.52 & 2.42 & 2.46 & 2.34 & {\bfseries 2.43} \\
   Human1 & 1.94 & 1.94 & 2.52 & 2.28 & {\bfseries 2.17} \\
   BC-Baseline & -0.43 & -0.23 & -0.19 & -0.42 & {\bfseries -0.32} \\
   Random & -1.80 & -1.29 & -1.14 & -1.16 & {\bfseries -1.35} \\
  \bottomrule
  \end{tabular}}
  {\caption{{\bfseries Leaderboard: normalized TrueSkill scores.} \smchange{The top three teams were GoUp, UniTeam, and voggite. GoUp achieved higher performance on all tasks but \cavetasknospace. We include scores for BC-Baseline (organizer-provided baseline), two expert humans, and a random agent.}\vspace{-8pt}}\label{tab:leaderboard}}
\end{table}

\subsection{Approaches of Competition Winners}
\label{subsec:winner_approaches}
\smchange{The scores of the top teams are captured in \Cref{tab:leaderboard}.}
\smchange{GoUp achieved the highest overall score of $2.09$, with UniTeam and voggite taking second- and third-place, respectively.}
\smchange{GoUp achieved the highest scores on all tasks except for \cavetasknospace. 
In this case, the UniTeam scored higher.
We now describe each team's solution in turn (more details in \Cref{sec:solutions}).} 

\begin{figure*}[t]
\centering
  \includegraphics[scale=0.4]{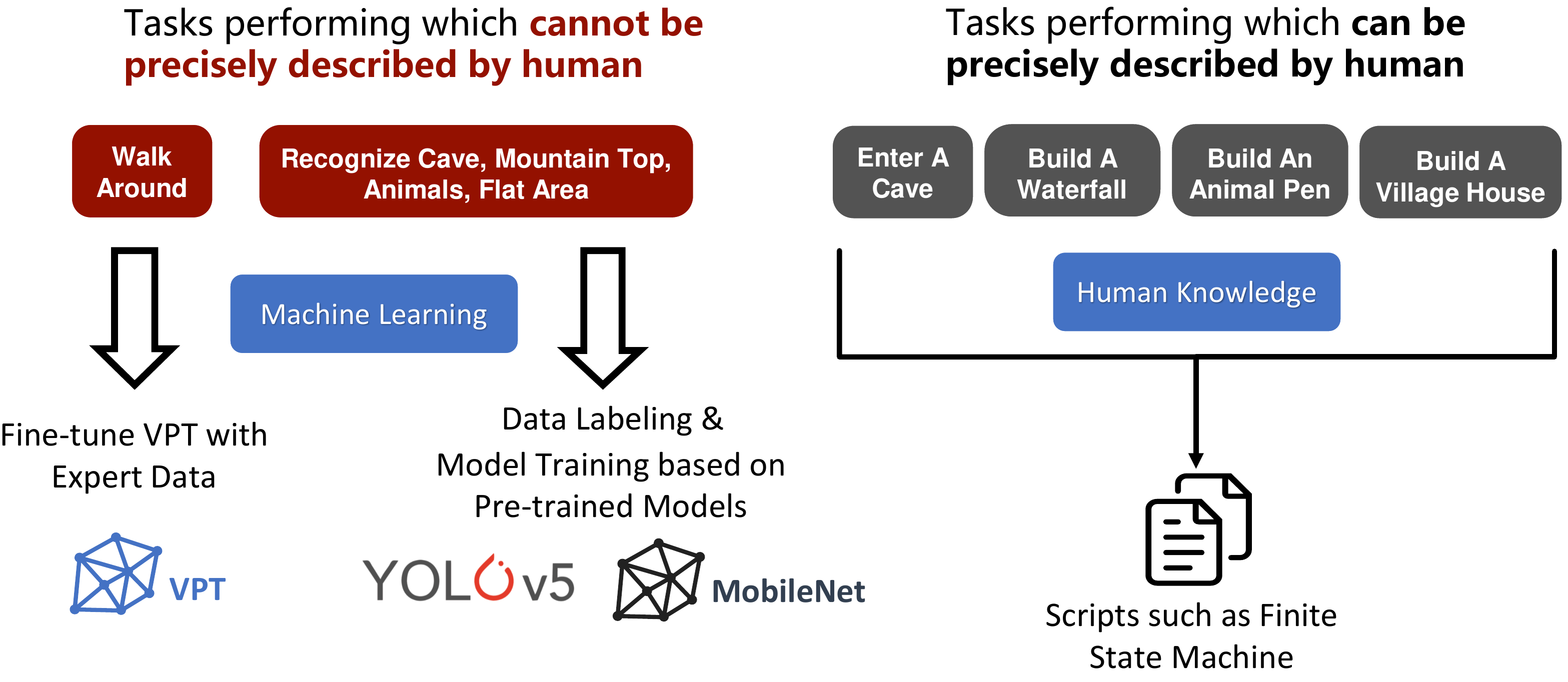}
  \caption{\smchange{\textbf{Decomposition of subtasks by GoUp.} To solve tasks that cannot be precisely described by a person, they leverage a variety of machine learning techniques. To solve the remaining tasks, they use human knowledge through scripting. \vspace{-8pt}} \label{Fig:Solution} }
\end{figure*}

\paragraph{First Place: GoUp}
\smchange{This solution utilized the power of machine learning and human knowledge by dividing each task into two parts: one that can be solved by transforming human knowledge into code (i.e., scripts) and the other that requires machine learning to solve.\footnote{Open-source code for GoUp: \url{https://github.com/gomiss/neurips-2022-minerl-basalt-competition}.} 
The team found that all of the four tasks consist of the same flow.
The agent walks around, searches for a target (e.g., a cave), then solves the task. 
They identified the targets in each task by training several classifiers and object detection models. 
To source data for training their models, they manually labeled images collected from the expert videos provided by the competition with the corresponding task. 
\Cref{Fig:Solution} shows the framework of the solution. 
Taking the AnimalPen task as an example, the solution for this task contains a fine-tuned VPT model for moving the agent, a fine-tuned YOLOv5 \citep{yolov5} detector for detecting the types and locations of animals, a fine-tuned MobileNet~\citep{howard2019searching} detector for identifying the location of fence placement, and a finite state machine that controls the executing flow of all the components. 
}

\paragraph{Second Place: UniTeam} 

\begin{figure}[t]
    \centering
    \includegraphics[width=\textwidth]{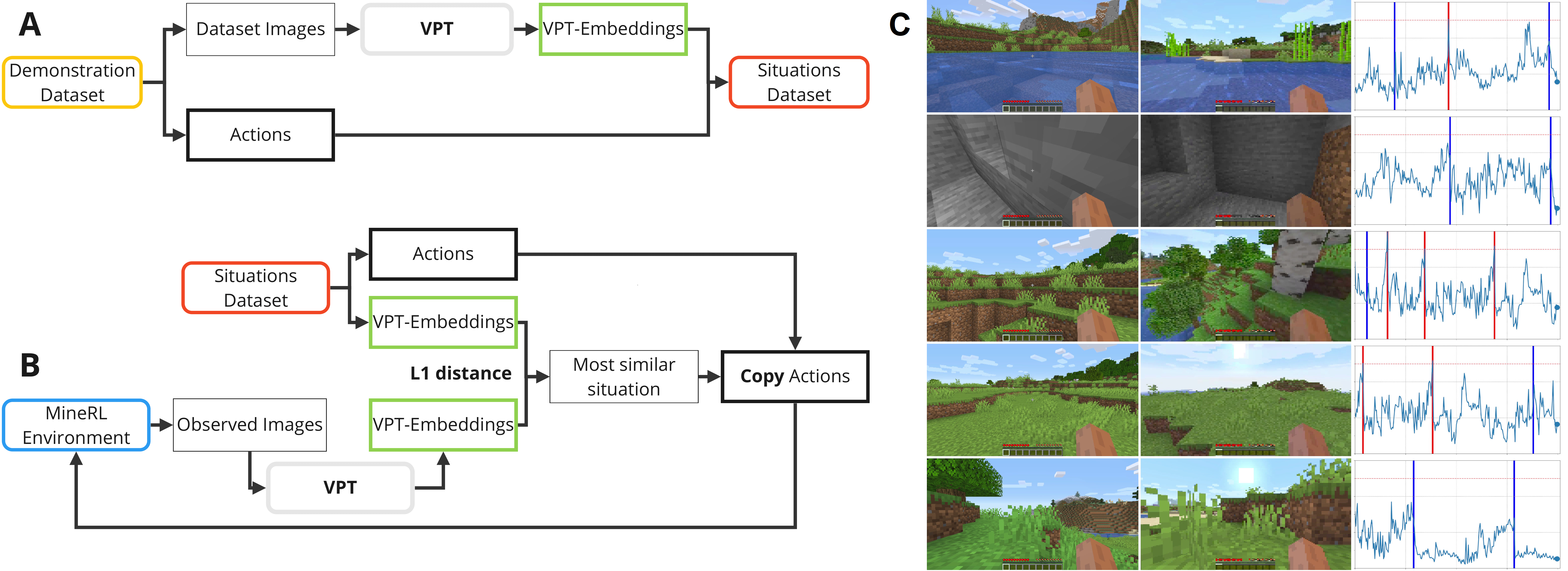}
    \caption{\smchange{\textbf{Generation of past-situation points in the VPT latent space by UniTeam.} (A) They encode situations from the BC dataset through a pre-trained VPT model. (B) They compute the L1 distance between their embedded current situation and the embedded situations from the expert’s dataset. Then, they copy actions from the best-matching reference situation. (C) Examples of visual (left) and numerical (right) similarity between current and reference situations. In the right figures, the blue vertical lines denote new time-based searches and red lines correspond to new threshold-based searches.
}\vspace{-8pt}}
    \label{fig:uniteam_approach}
\end{figure}

\smchange{UniTeam proposed a search-based behavioral cloning approach, which aims to reproduce an expert's behavior by copying relevant actions from relevant situations in the demonstration dataset.\footnote{Open-source code for UniTeam: \url{https://github.com/fmalato/basalt_2022_submission}.} 
\Cref{fig:uniteam_approach} shows their approach.
They defined a situation as a subset of consecutive frames and actions within a recorded trajectory, which they encoded using the VPT network.
To find the most similar situations to the current one, they used a pre-trained VPT model to produce latent representations.
They searched for the nearest embedding point in the VPT latent space to find the reference situation.
They assumed that retrieved situations represent an optimal solution to a specific past situation. 
As a result, they copy the corresponding actions.
}
\smchange{ 
After each timestep, they updated both the current and reference situations. 
Because the reference and current situations diverge over time, they measured the L1 similarity between the situations at each timestep. 
Once the distance between trajectories was greater than some fixed threshold (or after 128 timesteps), they performed a new search.
See~\cite{malato2022behavioral} for more details.}

\paragraph{Third Place: voggite} 
\smchange{This team focused on improving the behavioral cloning baseline.\footnote{Open-source code for voggite: \url{https://github.com/shuishida/minerl_2022}.}
To enable quick iterations of solutions, they precomputed the VPT state embeddings for all of the expert demonstrations, then trained a lightweight policy model with PyTorch Lightning. 
To focus on rarer but significant actions, they introduced action reweighting based on how frequently they were encountered. 
They observed that certain actions serve as signals that trigger higher-level changes in state. 
For example, the use of a bucket to pour water to complete the waterfall task signals that the agent should begin to climb down the mountain to take a scenic picture. 
They manually encoded these trigger actions along with the change in action distribution (e.g., decreasing the probability of moving forward once the agent starts building something); however, they plan to incorporate this idea into a hierarchical reinforcement learning framework, such as option-critic~\citep{bacon2017option}. 
}

\subsection{Approaches of Research Prize Winners}
\smchange{In addition to evaluating teams on the performance of their algorithms using human evaluations, we awarded additional prizes for research contributions.
To select the winners, each advisor read through a description of the approach and watched a video demonstration of the agent behavior. 
They had the option to view the agent's score. 
Each advisor then awarded a prize to the approach for their research contribution.
In general, the advisors preferred elegant, intuitive approaches that were ambitious, even if the final scores were relatively low. 
Advisors independently chose the following teams for the research prize: UniTeam (2 votes), KABasalt (2 votes), and KAIROS (1 vote). 
We now describe the remaining approaches in turn.
} 

\paragraph{KABasalt} 
\smchange{This team aimed for a reward modeling approach based on preference feedback after a pre-training phase through imitation and preference learning with demonstrations.\footnote{Open-source code for KABasalt: \url{https://github.com/BASALT-2022-Karlsruhe/ka-basalt-2022}.}
The main contribution was integrating the VPT models into the \textit{Imitation} library~\citep{gleave2022imitation} and {Stable-Baselines3}~\citep{stable-baselines3}, which required the ability to create compatible policy and reward model objects from the VPT backbone.
After tackling the myriad challenges of making reinforcement learning work with foundation models, such as incompatible interfaces, they are now further developing their solution and adding support for more features, like off-policy reinforcement learning algorithms.
}

\paragraph{KAIROS}
\begin{figure}[t]
    \centering
    \includegraphics[width=0.95\columnwidth]{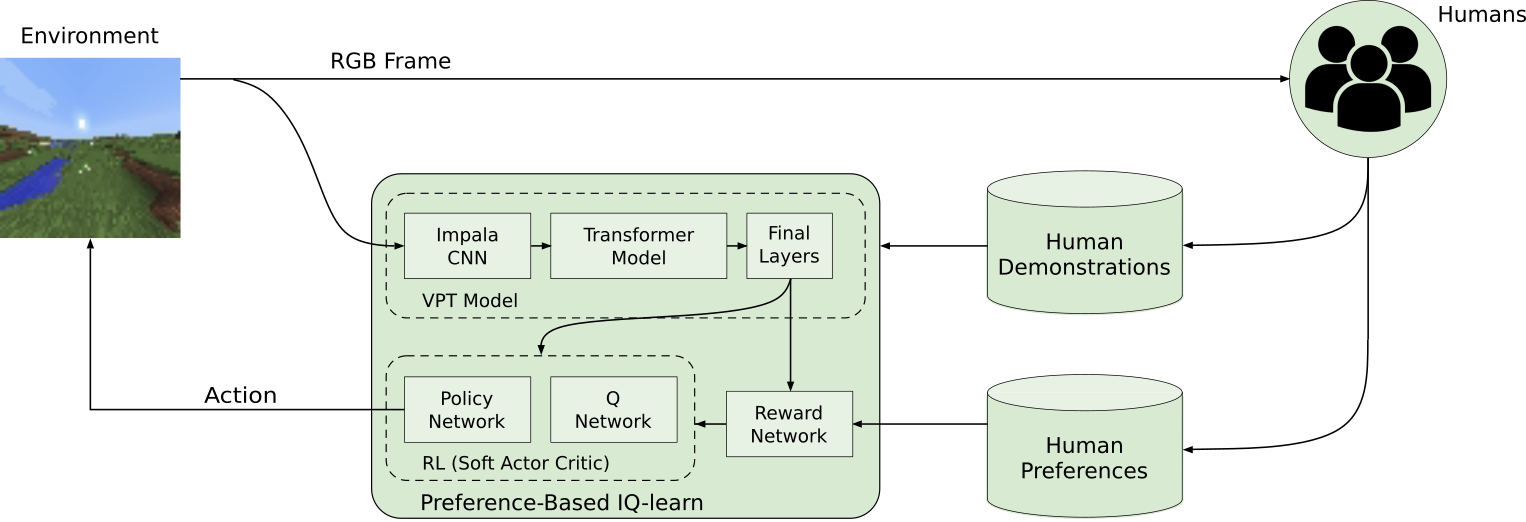}
    \caption{\smchange{\textbf{Preference-Based IQ-Learning (PIQL) algorithm proposed by KAIROS.} Their algorithm incorporates imitation learning with reinforcement learning using a reward model trained using pairwise human preferences over videos agents completing the tasks.}\vspace{-8pt}}
    \label{fig:piql_diagram}
\end{figure}

\smchange{This team proposed Preference-Based IQ-Learning (PIQL), a novel algorithm that extends a state-of-the-art actor-critic imitation learning algorithm (IQ-Learn~\citep{garg2021iq}) to additionally leverage the VPT model and online pairwise preferences over trajectories~\citep{christiano2017deep}.\footnote{Open-source code for KAIROS: \url{https://github.com/nwayt001/preference-IQL.git}.}
\Cref{fig:piql_diagram} shows their approach. 
PIQL uses pairwise preferences over videos of agents completing the task to learn a reward function via the Bradley-Terry model~\citep{christiano2017deep} through a reward head attached to the VPT network.
They define a novel critic loss that includes an IQ-learn term that prioritizes imitating the demonstrations and a reinforcement learning term~\citep{haarnoja2018soft} that aims to maximize the reward learned from human preferences. 
They include regularization terms that penalize the behavioral cloning loss on the demonstrations and the KL divergence between policies in successive learning iterations.
Thus, while performing imitation learning via IQ-Learn, PIQL obtains human pairwise preferences and learns a reward model that assists Q-function learning alongside the demonstrations.
}

\section{Discussion}
\label{sec:discussion}
\smchange{The results and outcomes of the competition indicate that there are parts that worked well and many opportunities for improvement.}

\paragraph{Task Design}
\kr{
While the four tasks stayed the same this year, the intended way to solve them changed. Fine-tuning the VPT model from human feedback meant that participants started with an agent already capable of traversing and exploring the Minecraft world. Compared to the last year, the solutions of the 2022 edition were indeed all able to navigate around the world, and they were mainly separable by the ability to place blocks down and coherency in their actions.
}
\smchange{We were again pleased to see that teams used significantly different approaches in tackling the tasks, which serves the research goals of the competition.}

\paragraph{Evaluation Methodology and Results}
This year, we used Amazon Mechanical MTurk to crowdsource the evaluation across multiple workers (see \Cref{app:evaluation_results} for details).
By using a scalable platform, we increased the number of human evaluators over the 2021 edition, which relied on a small pool of hired contractors.
To ensure high-quality responses, we set high qualification limits. 
We also asked the human judges to justify their choice of higher-performing agent, which we manually vetted to filter out low-quality answers. 
We found that asking for these responses helped substantially with filtering out low-quality answers.
We recommend utilizing this setup in future competitions that include human evaluations.

Compared to last year, we saw an increase in the performance of the submitted algorithms. 
Meanwhile, we also saw an increase in informative answers, where evaluators chose one of the two pairwise videos as better instead of answering ``draw". 
In 2021, we had a ``draw" rate of 80\%~\citep{shah2022retrospective}, while this year we had a 27-44\% ``draw" rate, depending on the task (see \Cref{app:evaluation_results}). 
\smchange{However, we note that this difference may be attributed to the change in crowd-sourcing platform from contract workers to MTurk. 
Regardless, this finding affirms the utility of this setup for assessing agents.
}




\paragraph{Intent of the Competition Compared with Evaluation Metrics}
\kr{When organizing an AI competition, one of the biggest questions is how to align the intent of the competition with the evaluation metrics. Especially in reinforcement or imitation learning competitions, there is usually a way to place high on the leaderboard without meaningfully following the intent. One way to solve this is through stringent rules on what methods are allowed, like in past MineRL Diamond competitions \citep{guss2019neurips}. The downside is the extra work required to check rule compliance and to continuously clarify the fine line between allowed and disallowed methods. The way we attempted to solve this tension was by allowing any methods but awarding extra prizes for interesting research contributions. The balance between the more objective evaluation metrics and the more subjective interestingness is one knob that could be tuned to achieve the goals of such competitions.} 

{An illustrative example of this tension is how team KAIROS decided to follow the intent of BASALT 2022.} 
\smchange{Although they did not score highly on the leaderboard, they received a research reward by closely following the intent of the competition.}
\smchange{In the future, to encourage creativity of solutions that adhere more closely to the intent of the competition, we recommend providing a diverse set of baselines for participants to build off of. 
Then, participants have examples of what the competition organizers are looking for and preliminary working code to further refine. }
\section{Conclusion}
\label{sec:conclusion}
\smchange{
We ran the MineRL BASALT Competition on Fine-Tuning from Human Feedback at NeurIPS 2022 to promote research on fine-tuning techniques that enable agents to accomplish tasks without crisply-defined reward functions. 
We described the competition, summarized the top solutions, and investigated the performance of the competitors. 
We believe that this competition achieved its main research goals of encouraging the development of algorithms to solve hard-to-specify tasks. 
However, we identified and provided concrete avenues for improvement in the future.
}

\acks{\smchange{
Running this competition was only possible with the help of many people and organizations.
FTX Future Fund, Microsoft, Encultured AI, and AI Journal provided financial support.
We thank our amazing advisory board: Fei Fang, Kiant\'e Brantley, Andrew Critch, Sam Devlin, and Oriol Vinyals for their advice and guidance. We thank Skylar Anastasia Ekamper and Martin Andrews for supporting other participants of the competition. 
We thank Matthew Rahtz for providing detailed feedback on a draft of this paper.
Finally, we thank AIcrowd for their help and the MTurk workers for their efforts in evaluating submissions.}}

\bibliography{bib}

\newpage
\appendix

\section{Task Details}\label{apd:comp_details}

In this section, we provide more information about the competition tasks and the associated human demonstration data.
For more details, we refer an interested reader to the competition proposal~\citep{kanervisto2022basalt}

\subsection{Data}

Similarly to last year, we provided the participants with a dataset of human demonstrations of each task. 
To produce these demonstrations, we used the same infrastructure as MineRL Diamond~\citep{guss2019neurips}. 
Each demonstration consists of a sequence of state-action pairs, called a trajectory, contiguously sampled at every Minecraft game tick.
There are 20 game ticks per second.
Each state is includes an RGB frame from the player's perspective, as well as a set of features from the game state. 
Each action consists of all keyboard and ``mouse'' interactions (change in view, pitch and yaw), in addition to a representation of the player GUI interactions. 

\subsection{Competition Tasks}
\ak{
Here we detail the four tasks participants had to solve with their submissions.
The following text is verbatim from the competition webpage, i.e., the same exact info participants saw.
}
\mybox{
\paragraph{Find Caves task}
\begin{itemize}
    \item Description: Look around for a cave. When you are inside one, press ESCAPE to end the minigame.
    \item Clarification: You are not allowed to dig down from the surface to find a cave.
    \item Starting conditions: Spawn in ``plains" biome.
    \item Timelimit: 3 minutes (3,600 steps)
\end{itemize}
}

\mybox{
\paragraph{Waterfall task}
\begin{itemize}
    \item Description: After spawning in a mountainous area with a water bucket and various
             tools, build a beautiful waterfall and then reposition yourself to
             ``take a scenic picture" of the same waterfall by pressing the ESCAPE
             key. Pressing the ESCAPE key also ends the episode.
    \item Starting conditions: Spawn in \texttt{extreme\_hills} biome. Start with a waterbucket,
                     cobblestone, a stone pickaxe and a stone shovel.
    \item Timelimit: 5 minutes (6,000 steps)
\end{itemize}
}

\mybox{
\paragraph{Village Animal Pen Task}
\begin{itemize}
    \item Description: After spawning in a village, build an animal pen next to one of
                 the houses in a village. Use your fence posts to build one animal
                 pen that contains at least two of the same animal. (You are only
                 allowed to pen chickens, cows, pigs, sheep or rabbits.) There
                 should be at least one gate that allows players to enter and exit
                 easily. The animal pen should not contain more than one type of animal. (You may kill any extra types of animals that accidentally got into the pen.) Don’t harm the village. Press the ESCAPE key to end the minigame.
    \item Clarifications: You may need to terraform the area around a house to build a pen. When we say not to harm the village, examples include taking animals from existing pens, damaging existing houses or farms, and attacking villagers. Animal pens must have a single type of animal: pigs, cows, sheep, chicken or rabbit.
    \item Technical clarification: The MineRL environment may spawn player to a snow biome, which does not contain animals. Organizers will ensure that the seeds used for the evaluation will spawn the player in villages with suitable animals available near the village.
    \item Starting conditions: Spawn near/in a village. Start with fences, fence gates, carrots, wheat seeds and wheat. This food can be used to attract animals.
    \item Timelimit: 5 minutes (6,000 steps)
\end{itemize}
}

\mybox{\
\paragraph{Village House Construction task}
\begin{itemize}
    \item Description: Taking advantage of the items in your inventory, build a new house in the style of the village (random biome), in an appropriate location (e.g. next to the path through the village), without harming the village in the process. Then give a brief tour of the house (i.e. spin around slowly such that all of the walls and the roof are visible). Press the ESCAPE key to end the minigame.
    \item Clarifications:  It’s okay to break items that you misplaced (e.g. use the stone pickaxe to break cobblestone blocks). You are allowed to craft new blocks. You don’t need to copy another house in the village exactly (in fact, we’re more interested in having slight deviations, while keeping the same “style”). You may need to terraform the area to make space for a new house. When we say not to harm the village, examples include taking animals from existing pens, damaging existing houses or farms, and attacking villagers. Please spend less than ten minutes constructing your house.
    \item Starting conditions: Spawn in/near a village (of any type!). Start with varying construction materials designed to cover different biomes.
    \item Timelimit: 12 minutes (14,400 steps)
\end{itemize}
}

\subsection{Intro Track Task (Obtain diamond shovel)}
\ak{In the \easy track, participants were challenged to obtain a diamond shovel. This is akin to the ``obtain diamond" task of the previous MineRL competitions, but the agent has to use the gathered diamond to craft a diamond shovel. While difficult, this task is easier with the OpenAI VPT models, one of which is able to obtain diamonds in 20\% of the trajectories.}

\ak{Upon submission, the agent was run for 20 episodes, starting from a fresh Minecraft world. Agent receives a reward for reaching a new stage in the crafting tree, much like in MineRL \citep{guss2019neurips}. The addition is that the agent receives a 2048 reward for crafting a diamond shovel. The final agent score is the maximum score over the 20 episodes. We used max instead of mean in this track, as its purpose was not to draw people to compete in it, but to introduce new users to the environment code and submission pipeline.}

\section{Participant Outcomes}
\label{sec:solutions}
Here we provide more details on participation and the solutions taken by the teams.

\subsection{Participation Details}
\begin{table}[t]
    \centering
    \begin{tabular}{ccccc}
    \toprule
         & 2019 & 2020 & 2021 & 2022 \\
         \midrule 
       Number of Messages & 2,473 & 2,464 & 7,047 & 5,800 \\ 
       Number of New Users  & 445 & 160 & 955 & 731 \\
       \bottomrule
    \end{tabular}
    \caption{\textbf{Discord statistics for the MineRL server from its inception in 2019 to 2022.} 
    These numbers indicate a substantial and sustained interest in using Minecraft for machine learning applications. }
    \label{tab:discord_stats}
\end{table}
We found that the increased interest mentioned earlier was also reflected in the MineRL Discord server.
We detail some interesting statistics in \Cref{tab:discord_stats}.
These numbers indicate a sustained interest in using Minecraft for machine learning research.

\subsection{Solution Details}
We now provide more details about some of the solutions.
\paragraph{GoMiss}
\begin{table}[t]
    \centering
    \begin{tabular}{ccc}
    \toprule 
     Recognition Target & Data Used & Data Type \\
    \midrule 
     Cave and hole & 10k images & Labeled with `Cave' and `Hole'  \\
     (object detection) & & \\ 
       Animal  & 4k images & Labeled with target animals \\ 
      (object detection) & & \\
       Flat area  & 10k images & 5k positive, 5k negative \\ 
      (classification) && \\
       Mountain top & 1750 images & 250 positive, 1500 negative \\
       (classification) && \\
      \bottomrule
    \end{tabular}
    \caption{\textbf{Data used by GoMiss to train the neural networks used to identify the targets for each task.} Interestingly, the most data was used to identify caves/holes and flat areas.}
    \label{tab:gomiss_data}
\end{table}
As mentioned in \Cref{subsec:winner_approaches}, GoUp utilized a number of classifiers and object detectors to recognize the targets for each task.
\Cref{tab:gomiss_data} contains details about the data used to train these neural networks.

GoUp finetuned the VPT model using imitation learning with the organizer-provided \cavetasknospace{} videos.
Notably, the fine-tuned Mobilenet was only used to identify the flat area. 
After finding a flat area, they used a script to verify whether a flat area was actually found.
At a high level, this script has the agents take actions, then checks the difference in the image observations.
The script also includes actions to help the agent make the ground more flat by digging or placing blocks. 
After all of these attempts, if the agent is not on a flat area, it will move to a new area to try again.

To build a house, GoUp implemented a series of hard-coded macro actions to construct each module.
Using the game image as input, they determine the current stage and execute the specific macro actions accordingly. 
These macro actions include: constructing pillars, walls, and doors. 
They did not use privileged state information, only the game image as input for the Hough transforms to accurately adjust the player's orientation.

\paragraph{KAIROS}
The videos used by their approach are rollouts generated by the agent during training.
The agent executes its current policy in the environment throughout training to generate rollout trajectories.
These trajectories are used in both preference-based RL and imitation learning via IQ-learn (which uses both expert demonstrations and on-policy data).
Overall, their approach could be characterized as combining IQ-learning from demonstrations, deep RL from human preferences, and a behavioral cloning term in the policy loss on a VPT torso. 
In PIQL, a Q-function is trained using both IQ-learn and deep RL from human preferences.

\section{Evaluation Process Details}
\label{app:evaluation_results}
\begin{figure}[t]
    \centering
    \includegraphics[scale=0.5]{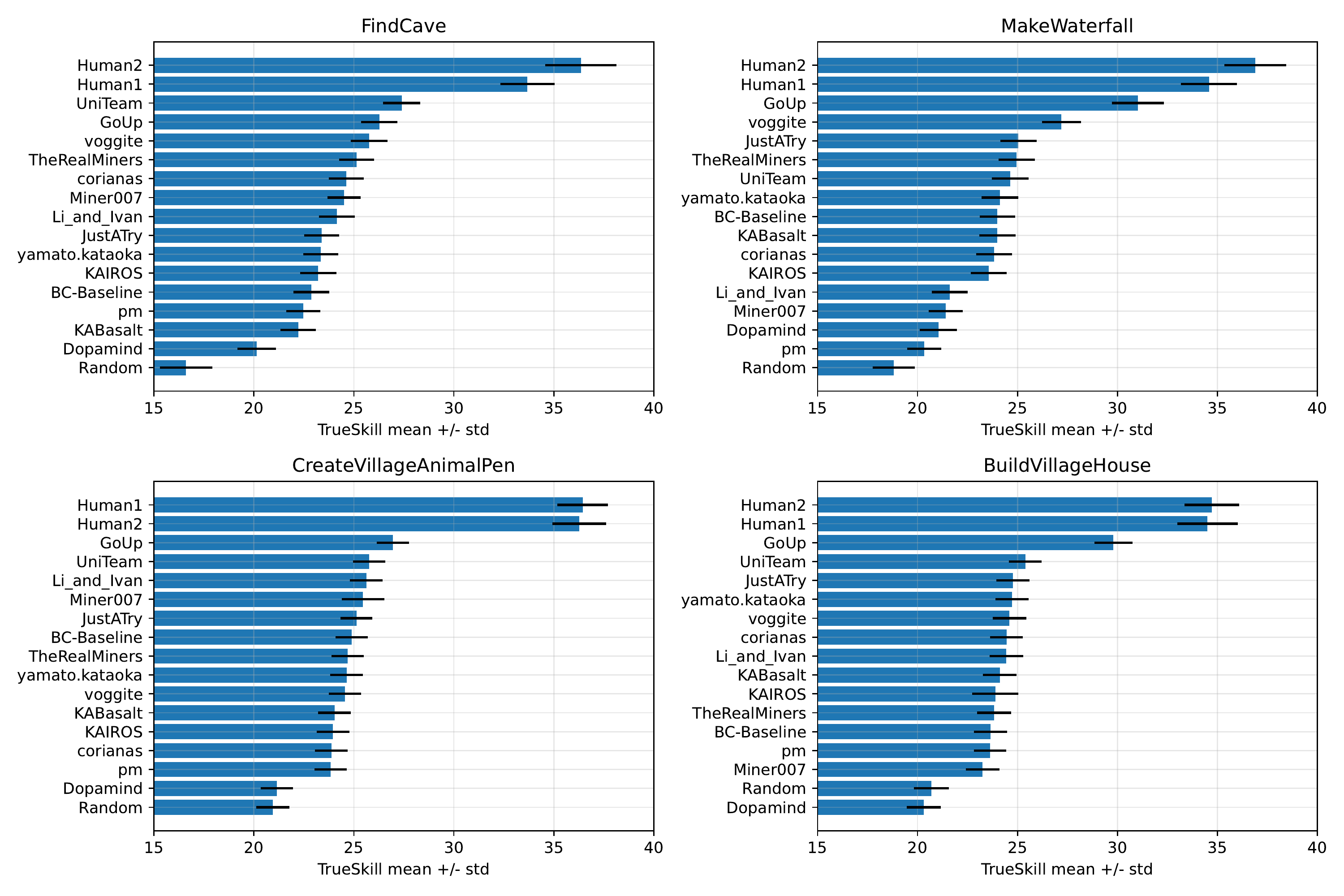}
    \caption{\textbf{TrueSkill rankings of the participants, including baselines.} The error bar shows plus/minus one standard deviation of the TrueSkill rating. ``Random" is an agent that chose actions randomly. ``BC\_baseline" is the behavioural cloning baseline supplied to the participants as a baseline solution.}
    \label{fig:evaluation_scores}
\end{figure}

\begin{figure}[t]
    \centering
    \includegraphics[scale=0.5]{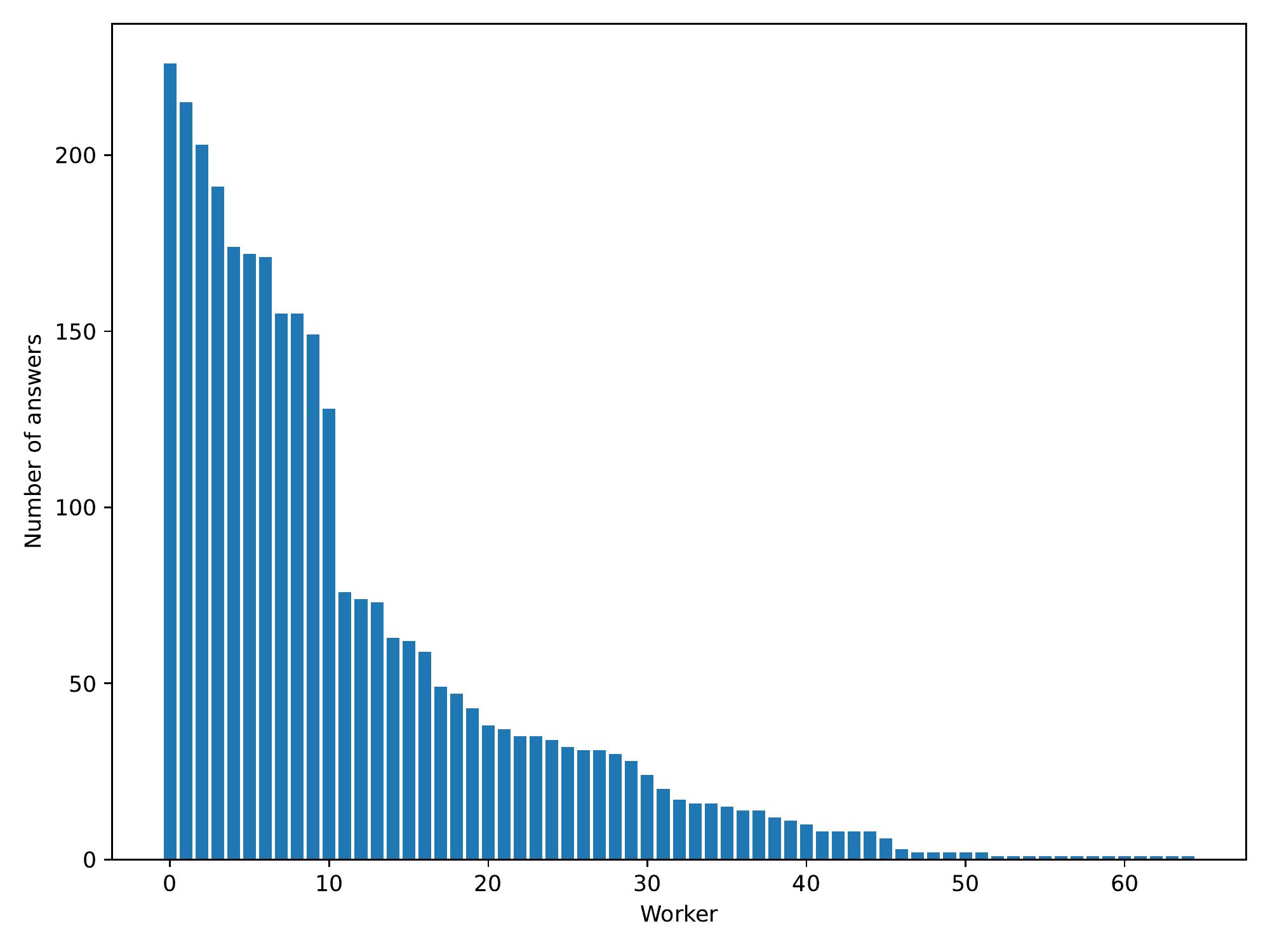}
    \caption{\textbf{Distribution of answers per human evaluator.} Here, the x axis represents a worker ID. Around 10 workers contributed over 100 responses each. The remaining around 50 workers contributed less than 100 responses each.}
    \label{fig:worker_distribution}
\end{figure}

\subsection{Final scores and MTurk worker statistics}
    Figure \ref{fig:evaluation_scores} shows the final TrueSkill rankings of each evaluated submission in each task. We also recorded two human players completing the same tasks in the same 10 evaluation seeds and a random agent to act as an upper and a lower bound of performance.

    In total, we had 3049 valid answers from 65 workers after removing 417 low-quality MTurk answers. We required the MTurk workers to have above 99\% HIT acceptance rate and more than 10,000 accepted HITs, as well as complete custom questionnaire testing for basic Minecraft knowledge. We then studied the quality of the mandatory justifications of answers to detect low quality answers (e.g., duplicates and irrelevant text) and remove them from the dataset.
     
    Figure \ref{fig:worker_distribution} shows the distribution of answers per worker, which shows that even the most active worker only submitted 10\% of the answers, indicating a healthy diversity in the worker pool.

\begin{table}[t]
\centering
\begin{tabular}{cccc}
\toprule
Task & Total & Draws & Draw \%\\
\midrule
FindCave & 722 & 201 & 27.84\% \\
MakeWaterfall & 682 & 210 & 30.79\% \\
CreateVillageAnimalPen & 914 & 404 & 44.20\% \\
BuildVillageHouse & 731 & 320 & 43.78\% \\
\bottomrule
\end{tabular}
\caption{\textbf{Comparison of the total number of \textit{draw} answers to total number of answers per task, across all submissions.} 
Draw means that the human evaluator marked both solutions as equally valid. 
Proportionally, the FindCave task had the least amount of \textit{draw} answers, while the CreateVillageAnimalPen task had the most.}
\label{tab:task_distribution}
\end{table}
    Table \ref{tab:task_distribution} shows the distribution of answers per task, as well as the amount of ``draw" votes. The increased numbers in \pentask and \housetask indicate the solutions were not as distinguishable from each other as in \cavetask and \waterfalltask. Given this increased amount of uninformative draw answers, we increased the number of answers for \pentask and \housetask to reach stabler results.

\subsection{TrueSkill normalization}
    We computed the final score as follows:
    \begin{enumerate}
        \item A TrueSkill rating consists of a mean and standard deviation. Take the mean parts of the TrueSkill ratings and normalize them per task to standard norm $\hat x = (x - \mu) / \text{max}(\sigma, 1)$, where $\mu$ and $\sigma$ are sample mean and standard deviation over the mean parts of TrueSkill ratings per task. (We clip the standard deviation to be at least 1 to avoid amplifying noise in rankings for tasks where all submissions behave nearly identically, where we would expect means to be similar and so the standard deviation would be tiny.)
        \item We then sum the normalized scores over tasks to form the final score.
    \end{enumerate}

\end{document}